\documentclass[sn-standardnature]{sn-jnl}

\theoremstyle{thmstyleone}%
\theoremstyle{thmstyletwo}%

\theoremstyle{thmstylethree}%
\usepackage{booktabs} 
\usepackage{hyperref} 
\usepackage{graphicx}
\usepackage{listings} 
\usepackage{textcomp}
\usepackage{siunitx}

\begin{document}

\title[ ]{EfficientBioAI: Making Bioimaging AI Models Efficient in Energy, Latency and Representation}


\author[1]{\fnm{Yu} \sur{Zhou}}

\author[1]{\fnm{Justin} \sur{Sonneck}}

\author[1]{\fnm{Sweta} \sur{Banerjee}}

\author[1]{\fnm{Stefanie} \sur{Dörr}} 

\author[1]{\fnm{Anika} \sur{Grüneboom}} 

\author[1,2]{\fnm{Kristina} \sur{Lorenz}}

\author*[1]{\fnm{Jianxu} \sur{Chen}}\email{jianxu.chen@isas.de}

\affil[1]{\orgname{Leibniz-Institut für Analytische Wissenschaften – ISAS – e.V.}, \orgaddress{\city{Dortmund}, \country{Germany}}}

\affil[2]{\orgname{Institute of Pharmacology and Toxicology, University of Würzburg},\orgaddress{\city{Würzburg}, \country{Germany}}}

\abstract{Artificial intelligence (AI) has been widely used in bioimage image analysis nowadays, but the efficiency of AI models, like the energy consumption and latency is not ignorable due to the growing model size and complexity, as well as the fast-growing analysis needs in modern biomedical studies. Like we can compress large images for efficient storage and sharing, we can also compress the AI models for efficient applications and deployment. In this work, we present EfficientBioAI, a plug-and-play toolbox that can compress given bioimaging AI models for them to run with significantly reduced energy cost and inference time on both CPU and GPU, without compromise on accuracy. In some cases, the prediction accuracy could even increase after compression, since the compression procedure could remove redundant information in the model representation and therefore reduce over-fitting. From four different bioimage analysis applications, we observed around 2-5 times speed-up during inference and 30-80$\%$ saving in energy. Cutting the runtime of large scale bioimage analysis from days to hours or getting a two-minutes bioimaging AI model inference done in near real-time will open new doors for method development and biomedical discoveries. We hope our toolbox will facilitate resource-constrained bioimaging AI and accelerate large-scale AI-based quantitative biological studies in an eco-friendly way, as well as stimulate further research on the efficiency of bioimaging AI. } 

\maketitle

Over the last decade, microscopy bioimaging techniques have been advancing at unprecedented pace, with higher spatial resolution~\cite{lelek_single-molecule_2021}, larger imaging volumes~\cite{yang_daxihigh-resolution_2022} and higher throughput for screening~\cite{wollman_high_2007}. These advancements have also led to the rapid development of artificial intelligence (AI) methods in microscopy image analysis tools (e.g. DeepImageJ~\cite{gomez-de-mariscal_deepimagej_2021}, ZeroCostDL4Mic~\cite{von_chamier_democratising_2021}, CellPose~\cite{pachitariu_cellpose_2022}, mmv\textunderscore im2im~\cite{sonneck_mmv_im2im_2023}). As the AI-based microscopy image analysis methods setting new records in various benchmarks and permitting quantitative biological studies not feasible before, we want to raise the awareness of another aspect of bioimaging AI models' performance: efficiency. Here, we consider efficiency from three different perspectives, latency efficiency, energy efficiency, and representation efficiency.
Generally, the complexity of the models (e.g., the number of layers and types of layers in a neural network) grew considerably in order to address more and more challenging problems. As a result, the network latency (i.e., the inference time) might increase significantly, especially when running on CPU or edge devices (e.g., running ``Smart Microscope" with an embedded system \cite{mahecic_event-driven_2022}) and on very large images. As a further consequence, the growing computation overhead led to substantial energy consumption. For example, the energy utilization for one forward pass with state-of-the-art AI models during inference has increased exponentially, rising from 0.1$J$ in 2012 to approximately 30$J$ in 2021~\cite{desislavov_trends_2023}. From a different perspective, in the context of bioimaging, the amount of training data with labels cannot grow at a commensurate pace as the complexity of the AI models, therefore the models themselves may become less efficiently trained and more likely to suffer from overfitting~\cite{xu_quantization_2018}.

One way to address the aforementioned issues would be ``compressing" the models to improve their efficiency, just like we compress large microscopy images for efficient storage. In the machine learning research literature, there are several potential solutions, such as neural network quantization, network pruning, and knowledge distillation. Quantization methods compress the models by quantizing the weights and activations of the layers into a lower precision (e.g., 8-bit integer) from commonly used floating points numbers (32-bit or 16-bit). Reduction on energy footprint and latency is achieved as the low-precision operations use much less energy and require much less memory access overhead than the original model. Network pruning aims to reduce the complexity of deep neural networks by setting the weights with low importance to zero, or removing the entire neuron if it is considered as not important at all. Finally, knowledge distillation is a compression strategy that transfers the knowledge of one or several complex ``teacher" models into a simpler and faster ``student" network. Due to model capacity gaps between the student and teacher models, knowledge distillation is more suitable for multi-teacher simplification or cross-modality distillation rather than regular model compression ~\cite{gou_knowledge_2021}.
    



Even though model compression has been widely applied in the computer vision field, such as in autonomous driving~\cite{ding_req-yolo_2019}, to the best of our knowledge, these techniques and tools (e.g.,~\cite{kozlov_neural_2020,li_mqbench_2021,siddegowda_neural_2022,microsoft_neural_2022}) are still not easily accessible in the field of bioimaging AI for several reasons. Firstly, they are not designed specifically for biologists, so it is nearly impossible to apply them in common AI-based bioimage analysis tools. Meanwhile, most of them only supports specific backends (e.g. nncf~\cite{kozlov_neural_2020} for intel platform and AIMET~\cite{siddegowda_neural_2022} for Qualcomm chips), which hinders the general compatibility with existing bioimaging AI models. Furthermore, at present the compression techniques have only been applied to biomedical imaging AI models in simulated settings, as their deployment in practice to achieve real acceleration and energy reduction remains elusive~\cite{xu_quantization_2018,askarihemmat_u-net_2019}. 
    \begin{figure}[t]
        \centering
        \includegraphics[width=\linewidth]{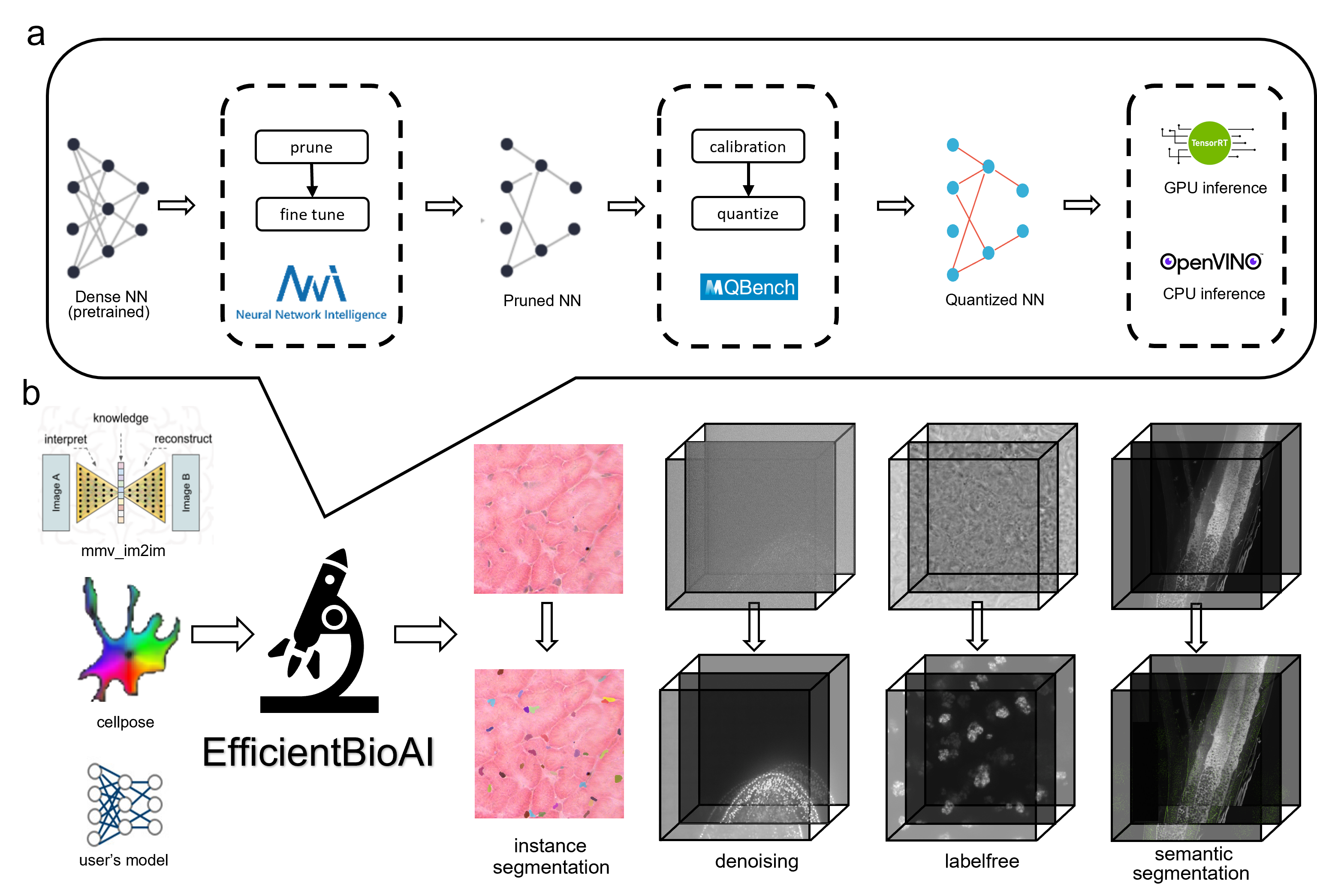}
        \label{fig:a}
        \caption{\textbf{overview of the toolbox.} EfficientBioAI aims to compress the model and accelerate the bioimage tasks. \textbf{a.}  Given a pretrained neural network (NN) in pytorch, the whole pipeline consists of two phases: compression (pruning, quantization) and inference. The compressed models are able to run on different hardwares with specific inference engine. Meanwhile, various compression strategies and algorithms are off-the-shelf and can be chained up to build the compression pipeline. \textbf{b.} Different types of models are supported and deployable. Users can also deploy their own models. The results are justified effective on multiple bioimage analysis tasks.}
        \label{fig:Method_train}
    \end{figure}
    
To address this problem, in this project we introduce a ``plug-and-play" toolbox called EfficientBioAI, aiming to bring state-of-the-art model compression techniques accessible for the bioimaging AI community. ``Plug-and-play" is achieved from two aspects. First, the toolbox is released as a package on PyPI, which allows users to install it effortlessly, obviating the need for setting up intricate environmental configurations. Second, the toolbox is designed to work off-the-shelf without changing users' bioimaging AI codebase, as long as the code is in PyTorch and contains no ``dynamic control flow" in model configuration (see Supplement Methods). While maintaining its simplicity for easy-use by default, the toolbox also provides flexibility to suit different needs, such as easy switch between CPU and GPU, customizable compression levels, etc. 

The overall concept of our tool is illustrated in Fig \ref{fig:Method_train}. The entire workflow can be divided into two phases: compression and inference. In the compression phase, the tool receives a pretrained model and compresses it using specific algorithms with specific level of compression. For example, users can choose the sparsity ratio in the pruning step or even skip the pruning step (sparsity ratio = 0) and go to quantization directly (larger pruning ratio may have higher level of compression and therefore more latency and energy footprint reduction, but more likely to negatively affect model accuracy). Fine-tuning and calibration techniques are automatically employed to avoid accuracy degradation. During the inference stage, users can choose to deploy the compressed model on either CPU or GPU, where a proper inference engine (e.g., OpenVINO or TensorRT) will be automatically triggered to achieve the best performance. 

We evaluated the efficiency gain by using the EfficientBioAI toolbox on four different bioimage analysis problems with both CPU and GPU: 3D semantic segmentation of osteocyte lacunae from lightsheet fluorescence microscopy images of mouse cortical bones, 3D labelfree prediction of fluorescent images of fibrillarin from transmitted light images of hiPS cells, 3D denoising of microscopy images of Panaria, and 2D instance segmentation of nuclei from whole-slide images (WSI). All three 3D problems were tested with our open-soruce package for microscopy image-to-image transformation toolbox, mmv\_im2im~\cite{sonneck_mmv_im2im_2023}, considering its general applicability on all kinds of microscopy image-to-image transformations. Meanwhile, to demonstrate the compatibility, the 2D nuclei segmentation problem was tested with CellPose \cite{pachitariu_cellpose_2022}. Additional demonstration on how to use our toolbox with other bioimaging AI tools (e.g., ZeroCostDL4Mic \cite{von_chamier_democratising_2021}) in a plug-and-play manner can be found as tutorials on our Github page.

\begin{table*}[t]
\centering
\caption{Compression performance on multiple tasks with different hardware}
\label{tab:quantization_perf}
\resizebox{\linewidth}{!}{
    \begin{tabular}{cccllcc}
       \toprule
       task & method & hardware & latency(s) & energy(10$^{-3}$kWh) & accuracy* & image size \\
       \midrule
                     & fp32 & CPU & 14.05 & 0.814 & 0.770 &\\
       & int8 & CPU & 7.46 & 0.387(52.5$\%$ $\downarrow$) & 0.798 & \\
       3D denoising    & pruning+int8 & CPU & 7.42 & 0.384(52.8$\%$ $\downarrow$) & 0.811 & (95$\times$1024$\times$1024)\\
       & fp32 & GPU & 4.73 & 0.280 & 0.806 &\\
                     & int8 & GPU & 1.57 & 0.091(67.4$\%$ $\downarrow$) & 0.811 &\\
                     & pruning+int8 & GPU & 1.21 & 0.085(69.6$\%$ $\downarrow$) & 0.815 &\\
       \midrule
                        & fp32 & CPU & 383.04 & 18.044 & 0.581 &  \\
                        & int8 & CPU & 238.12 & 9.256(48.7$\%$ $\downarrow$) & 0.565 &  \\ 
         2D instance seg & pruning+int8 & CPU & 196.06 & 6.840(62.1$\%$ $\downarrow$) &0.554 &(11578$\times$11660)\\
       & fp32 & GPU & 129.06 & 5.677 & 0.581 & \\
                        & int8 & GPU & 104.31 & 4.057(28.5$\%$ $\downarrow$) &  0.565 &\\
                        & pruning+int8 & GPU & 97.05 & 3.650(35.7$\%$ $\downarrow$)  & 0.567 &\\
       \midrule
                         & fp32 & CPU & 32.28 & 1.922 & 0.865  &\\
          & int8 & CPU &  24.46 & 1.316(31.5$\%$ $\downarrow$)  & 0.864 & \\
       3D labelfree   & pruning+int8 & CPU & 24.29 & 1.307(32.0$\%$ $\downarrow$) &  0.791 & (65$\times$924$\times$624)\\
        & fp32 & GPU & 8.63   & 0.526 & 0.864 &\\
                         & int8 & GPU & 1.97    & 0.115(78.1$\%$ $\downarrow$) & 0.864 &\\  
                         & pruning+int8 & GPU & 1.645  & 0.102(80.6$\%$ $\downarrow$)  & 0.797 & \\
        \midrule
                         & fp32 & CPU & N/A & N/A & N/A  &\\
          & int8 & CPU &  9608.61 & 485.442  & 0.656 & \\
       3D semantic seg   & pruning+int8 & CPU &  9383.27 & 473.620  & 0.685 & (297$\times$2048$\times$2048)\\
        & fp32 & GPU & 350.01   & 97.272 & 0.667 &\\
                         & int8 & GPU & 303.27    & 85.087(12.5$\%$ $\downarrow$) & 0.673 &\\   
                         & pruning+int8 & GPU & 285.12    & 81.744(16.0$\%$ $\downarrow$) & 0.687 &\\ 
       \bottomrule
       \multicolumn{7}{l}{\footnotesize *Pearson correlation coefficient for denoising and labelfree task, ap50 for instance segmentation task, dice for semantic segmentation task. }\\
    \end{tabular}
}
\end{table*}

We measured the efficiency on these tasks with fp32 precision (original models), int8 precision and pruning+int8 compression strategy, utilizing two different inference engines: OpenVINO for CPU and TensorRT for GPU. The metrics employed to evaluate performance on both original models and compressed models were Pearson correlation for the labelfree and denoising tasks, average precision at 50\% IoU (ap50) for the 2D instance segmentation task, and Dice coefficient (Dice) for the 3D semantic segmentation task. In addition, we reported latency in seconds (s) and energy consumption in watt-hours (Wh) using CodeCarbon\footnote{\url{https://github.com/mlco2/codecarbon}}. All results are shown in Table \ref{tab:quantization_perf}.

For the 3D denoising task, on CPU end we observed a slight increase in Pearson correlation, with a 1.9$\times$ speedup and a 52.5$\%$ energy footprint reduction. The performance will be slightly improved if the pruning technique is applied. Notably, the observed rise in the metric could be attributed to the enhancement of model representation efficiency through the compression step, resulting in a decrease in overfitting~\cite{xu_quantization_2018}. Meanwhile, for the 3D labelfree problem, the accuracy kept almost unchanged and a 4.4-fold acceleration and a 78.1$\%$ reduction in energy consumption were reported on GPU. In terms of the 3D semantic segmentation task, a 12.5\% energy drop was observed on GPU without accuracy lost. There is also some accuracy improvement if the pruning technique was added. The latency and energy of the original fp32 model were not recorded on CPU since we found it infeasible to complete within reasonable time in practice. Lastly, the latency and energy consumption also showed substantial improvements in the 2D instance segmentation task with negligible accuracy drop. We can have another 10\% energy drop if pruning technique is applied. 


In addition, we compared the influence of different compression algorithms. In the context of the pruning algorithm, different pruners and sparsity ratios are compared on the instance segmentation task (Supplementary Fig. \ref{fig:2} ). We found that the structured L1Pruner with a sparsity ratio of 50\% resulted in a minor accuracy loss, but provided significant energy savings and improved latency speed, thereby validating its effectiveness. 

The presented toolbox exhibits certain limitations. At present, it is unable to support several hardware, such as the Apple M-series chip and AMD platforms, which may marginally impede its application in bioimage AI tasks. Additionally, this tool solely enhances efficiency during the inference phase when provided with a pre-trained model. Its quantization function lacks the capacity to augment efficiency during the training phase, as gradients cannot be accurately estimated with low precision in the backpropagation step. Considering the successful application of human-in-the-loop techniques in the bioimaging AI model training phase, enhancing efficiency in the repetitive fine-tuning process would be beneficial. The emerging trend involves the utilization of fp8 for both training and inference phases, which not only improves training efficiency but also obviates the need for quantization in the inference step~\cite{micikevicius_fp8_2022}.

The proposed method holds great potential in the bioimaging AI field. Primarily, it is characterized by its user-friendliness and high extensibility, making it accessible to non-expert users who can benefit from the advantages of model compression. Notably, if 100 biologists were to process the Jump Target ORF dataset\footnote{\url{https://app.springscience.com/workspace/jump-cp}} (nearly 1 million images) using AI model reported before, our toolkit could save energy equivalent to the emissions generated by a car driven for roughly 7300 miles. More broadly, it is the very first attempt to focus on the efficiency of bioimaging neural networks and have the actual improvement in the latency and energy footprint.

In summary, our experiments provide evidence that employing diverse compression techniques can yield substantial enhancements in latency and energy efficiency across all tasks, while incurring only minimal or negligible performance degradation. These findings strongly support the potential of quantization and pruning as promising methodologies for deploying bioimaging AI models.

\section{Acknowledgment}
Y. Z., J.S., S.B. and J.C. are funded by the Federal Ministry of Education and Research (Bundesministerium für Bildung und Forschung, BMBF) in Germany under the funding reference 161L0272. All co-authors are also supported by the Ministry of Culture and Science of the State of North Rhine-Westphalia (Ministerium für Kultur und Wissenschaft des Landes Nordrhein-Westfalen, MKW NRW)

\bibliography{efficientbioai}
    
\section*{Data Availability}
The data will be uploaded to zonedo once upon approval.

\section*{Code Availability}
The whole project including the source code can be accessed from \url{https://github.com/MMV-Lab/EfficientBioAI}.

\newpage
\section*{Methods}

\subsection*{Quantization Algorithm}

Quantization function was achieved with the aid of MQBench~\cite{li_mqbench_2021} python package (version: 0.0.7)\footnote[1]{\url{https://github.com/ModelTC/MQBench}}, which characterizes itself by integrating state-of-the-art quantization algorithms. Post Training Quantization (PTQ) technique was applied in the compression pipeline. This method doesn't necessitate any retraining step, and only employed a calibration step to restore the precision. To do this, we randomly selected 5 samples in the evaluation set for the calibration. During the process of calibration, we chose EMAQuantileObserver to record the quantization information, such as the scaling value. In terms of the granularity, per-tensor quantization was chosen, meaning that the weight/activation in same layer share the same quantization information. Symmetric quantization strategy was applied and the precision after quantization is integer 8 (int8).

\subsection*{Prune Algorithm}

\begin{figure}[t]
    \centering
    \includegraphics[width=\linewidth]{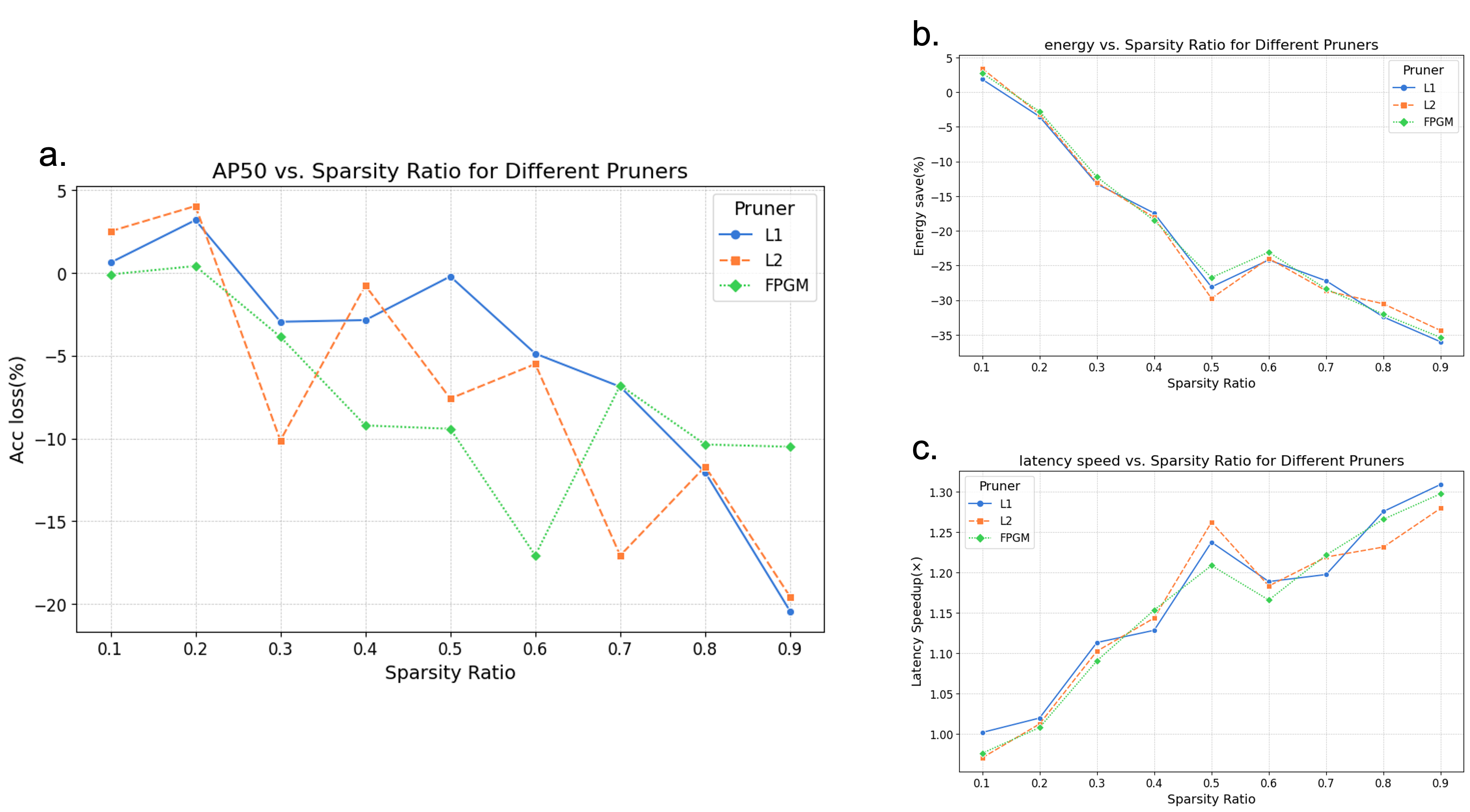}
    \label{fig:a}
    \caption{\textbf{The effect of the sparsity ratio on different pruners, i.e., pruning algorithms (L1, L2, and FPGM) by observing the ratio of performance degradation (a.), energy saving (b.) and latency speedup (c.).}}
    \label{fig:2}
    \end{figure}

The pruning algorithm is implemented on the basis of neural network intelligence (nni) package (version 2.1), which is capable of optimizing neural network based on multiple AutoML strategies \footnote[2]{\url{https://github.com/microsoft/nni}}. Pruning can be customized by the user in the configuration yaml file, including the algorithm, granularity, criteria, sparsity ratio, etc. We used our cellpose 2d instance segmentation task as an example to choose the pruning algorithm and sparsity ratio. The results were shown on Fig. \ref{fig:2}. In specific, we evaluated the efficiency performance (accuracy, energy, latency) on three structured pruners (L1~\cite{han_learning_2015}, L2~\cite{han_learning_2015}, FPGM~\cite{he_filter_2018}) under different sparsity ratios. The result showed that L1 pruner with 50\% sparsity ratio can avoid significant performance degradation and achieve reasonable energy reduction and latency speedup. After pruning, the workflow conducted a model fine-tuning step by 1000 epochs to restore the accuracy. Meanwhile, hardware speedup was achieved by applying the SpeedUp tool provided by the nni package~\cite{microsoft_neural_2022}.

\subsection*{Inference Engine}

TensorRT (version 8.5.3.1) and OpenVINO (version 2022.3.0) were chosen for Nvidia GPU and Intel CPU inference back-ends, respectively. In the experiments reported in this paper, the type of CPU was Intel (R) Xeon (R) W-2245 CPU @ 3.90GHz and the type of GPU was Quadro RTX 4000. The dynamic batch inference was also enabled on both back-ends to ensure the arbitrary batch size of the input. This is crucial in the cellpose/omnipose model because the batch size of the input data could be different. The inference mode for different back-ends varies at the same time. In terms of TensorRT, asynchronous inference was executed while synchronous inference was used for OpenVINO. We didn't observe obvious performance difference between infer strategies in our experiment. \\

\subsection*{EfficientBioAI Toolbox Codebase}
The code structure contains two main parts, compression and inference. For the compression part, there is a compress.py script responsible for quantizing/pruning the pretrained PyTorch model using strategies specified by the user. During this process, a parser inherited from the BaseParser class will be instantiated to read in both the model and data. Then compression pipeline are applied to the model. The compressed model will be exported in the format of onnx. Finally the exported model will be transformed to the file format compatible with various back-ends. In terms of GPU end, we write a script called onnx2trt.py to switch the onnx file to the serialized TensorRT engine (.trt file). While model optimizer (mo) toolkit\footnote{\url{https://docs.openvino.ai/2022.3/openvino_docs_MO_DG_Deep_Learning_Model_Optimizer_DevGuide.html}} is employed to adapt the onnx model to the format suitable for the OpenVINO CPU inference (.xml .bin file).

A script called inference.py is in charge of the inference part. Multiple infer objects are called to run different inference tasks. Also there are two backend classes (OPVModule, TRTModule) for inferring on CPU/GPU hardware. After the inference, the evaluation metrics are collected. To extend the toolbox to user-specific special models or tasks, one may need to inherit the BaseParser and BaseInfer classes and overwrite some IO functions such as how to read their model and data. 

    \subsection*{User Guide}
    Full documentation and tutorials can be found at \url{https://github.com/MMV-Lab/EfficientBioAI}.
    \subsubsection*{Installation}
    \indent Both pip and docker installation are supported.
    \begin{itemize}
        \item pip
            \begin{enumerate}
                \item create the virtual environment:
                    \begin{lstlisting}
conda config --add channels conda-forge
conda create -n efficientbioai python=3.8 setuptools=59.5.0
                    \end{lstlisting}
                \item install the dependencies:
                    \begin{lstlisting}
git clone git@github.com:ModelTC/MQBench.git
cd MQBench
python setup.py install
cd ..
                    \end{lstlisting}
                \item then install the efficientbioai package:
                    \begin{lstlisting}
git clone git@github.com:MMV-Lab/EfficientBioAI.git
cd EfficientBioAI
pip install -e .[cpu/gpu/all] # for intel cpu, nvidia gpu or both
                    \end{lstlisting}
            \end{enumerate}
        \item docker
            \begin{enumerate}
                \item CPU:
                    \begin{lstlisting}
cd docker/cpu
bash install.sh # if not install docker, run this command first
bash build_docker.sh # build the docker image
cd ../..
bash docker/cpu/run_container.sh #run the docker container
                    \end{lstlisting}
                \item GPU:
                    \begin{lstlisting}
cd docker/gpu
bash install.sh # if not install docker, run this command first
bash build_docker.sh # build the docker image
cd ../..
bash docker/gpu/run_container.sh #run the docker container
                    \end{lstlisting}
            \end{enumerate}
    \end{itemize}

    \subsubsection*{Run}
    \indent They are two scripts for compression and inference, respectively. 
        \begin{itemize}
            \item Compression:
                \begin{lstlisting}
python efficientbioai/compress.py --config path/to/config --exp_path path/to/save                    
                \end{lstlisting}
            \item Inference:
                \begin{lstlisting}
python efficientbioai/inference.py --config path/to/config
                \end{lstlisting}

        \end{itemize}

\subsection*{Specification for off-the-shelf deployment}

The toolbox can be used off-the-shelf and serve theoretically any third-party PyTorch-based model as long as the modules inside the model doesn't contain any so-called ``dynamic control flow". For example, the hyperparameters of each layer such as kernel size should not be determined by the input of the data. Control flow (e.g., if/else) should also remain fixed during inference. If the PyTorch module is standard and subject to the  aforementioned rules, the whole pipeline is off-the-shelf. 

For example, the original CellPose unfortunately contains dynamic control flow. So, we employed the model by forking the Github repo and removing such control flow. For a model satisfying aforementioned requirement, e.g., a third-part U-Net based denoising model open-sourced on Github (\url{https://github.com/juglab/DecoNoising}), users only need to add a few lines of code to complete the compression. See full tutorial at \url{https://github.com/MMV-Lab/EfficientBioAI/tree/main/tutorial}.

\section*{Experiment Details}

We selected four typical tasks for demonstration, namely 3D microscopy image denoising, 2D nuclei instance segmentation, 3D semantic segmentation and 3D labelfree prediction. Experiment details are elaborated below.

\subsubsection*{3D denoising of planaria microscopic images}
    
In the context of microscopy imaging, There is always a trade-off between the imaging speed and signal-to-noise ratio (SNR) due to the hardware limitation~\cite{weigert_content-aware_2018}, so we could get low-resolution images if imaging speed has the higher priority. In this experiment we applied AI models to increase the SNR of the images, in order to maintain the image quality for the follow-up analysis. 

The dataset that we used was the denoising planaria dataset~\cite{weigert_content-aware_2018}. The model we chose here was a UNet based network provided by the monai package~\cite{cardoso_monai_2022}. 

\begin{figure}[tbh]
    \centering
    \includegraphics[width=\linewidth]{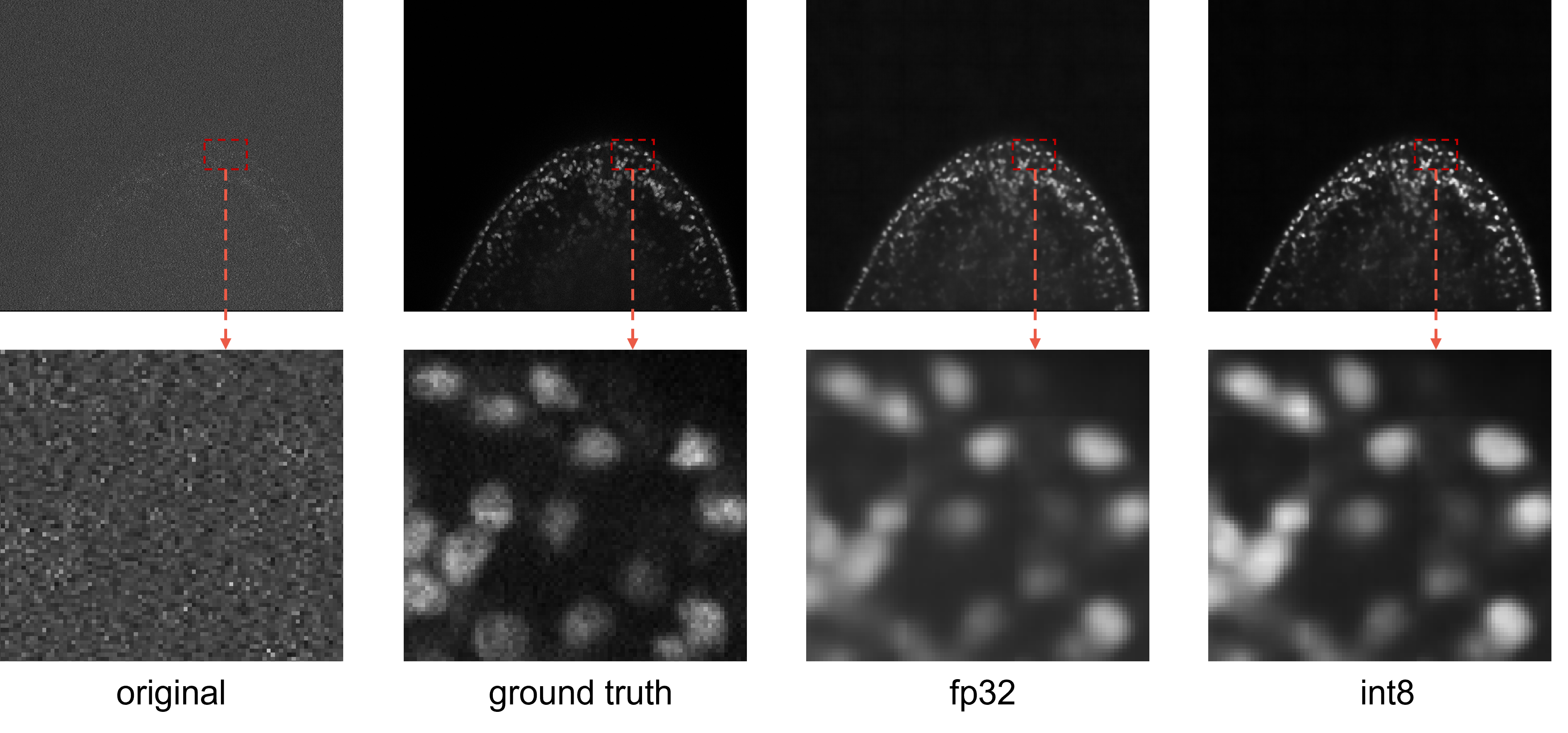}
    
    \caption{\textbf{Denoising Experiment Results.} 
     \textbf{Top row:} original image, ground truth, inference output with float32 precision, inference output with integer 8 precision \textbf{Bottom row:} The corresponding zoomed-in images. The focus of this paper is not to achieve the best denoising performance. Instead, the key is to make sure the prediction from the compressed model (int8) is close enough to the prediction from the original model (fp32).}
    \label{fig:3_denoise}
\end{figure}

For the inference step, totally 20 images were utilized. We used the sliding window strategy with 0.1 overlap and [32,128,128] as the window size. Example result images were shown in Fig. \ref{fig:3_denoise}.

\subsubsection*{2D nuclei instance segmentation with cellpose}

Histological analysis is the gold standard for the evaluation of tissue in biomedical research and it is applied for diagnostic and research purposes to quantify alterations in the diseased tissue or to qualitatively evaluate the pathological status. Different staining allows the visualization of certain cell types, tissues or subcellular compartments for example nuclei or lipid depositions. In cardiovascular research, the evaluation of cardiomyocyte size can give information about hypertrophic growth, whilst other staining can also visualize multi-nucleation or fibrotic areas. A commonly used staining is the hematoxylin/eosin (HE) staining. Hematoxylin stains basophilic structures such as DNA and therefore visualizes nuclei, while eosin binds to alkaline structures such as cytoplasmic protein and therefore visualizes the cytoplasm~\cite{cardiff_manual_2014}. HE staining is a common staining for the analysis of cardiomyocyte size. However, the analysis is time consuming, and the automatization is afflicted with errors. Cell borders need to be determined precisely, and the correct orientation of the cutting plane through the respective cardiomyocytes needs to be assessed, i.e. the plane needs to be vertical to exclude incorrect cell sizes because of semi-longitudinal planes. The latter can be ruled out by a rather central localization of the nucleus within the cardiomyocytes, and by a round rather than an oval or longitudinal shape of the cardiomyocytes.

Here, our task is to do 2D nuclei instance segmentation from HE stained images. The data were prepared and imaged as below: 

First, hearts of C57BL/6J mice were collected and fresh frozen in liquid nitrogen. Frozen heart tissue was embedded in Tissue-Tek\textregistered O.C.T.\texttrademark Compound (SA62550, Science Services) and placed on a cryostat cryotome plate. Tissue sections of 6 $\mu m$ were cut at \SI{-20}{\degreeCelsius}. Slides with tissue sections were stored at  \SI{80}{\degreeCelsius}. For HE staining, slides were left to dry at room temperature and stained directly thereafter. 

Then for cardiomyocyte size determination, tissue sections were stained with hematoxylin/eosin solution. Staining was performed with the ST5020 automatic staining machine (Leica) using a ready-made hematoxylin solution (MHS32, Sigma Aldrich Co.) and an eosin solution containing 1\% eosin (sigma Aldrich Co.) and 0.0125\% pure acetic acid (3738, Carl Roth GmbH \& Co. KG). Staining protocol was followed as previously described~\cite{tomasovic_interference_2020,lorenz_new_2009}. Stained slides were sealed with Cytoseal\texttrademark XYL (Thermo Fisher Scientific) and covered with a glass coverslip. Image acquisition was taken at 200$\times$ magnification with Zeiss Axio Scan.Z1. 

The model applied here is an in-house pretrained cellpose model. Cellpose is one of the mostly widely used bioimage segmentation tools. This is meant to demonstrate our compatibility and efficiency improvement on common third-party bioimaging AI tools.

\begin{figure}[!tbp]
    \centering
    \includegraphics[width=\linewidth]{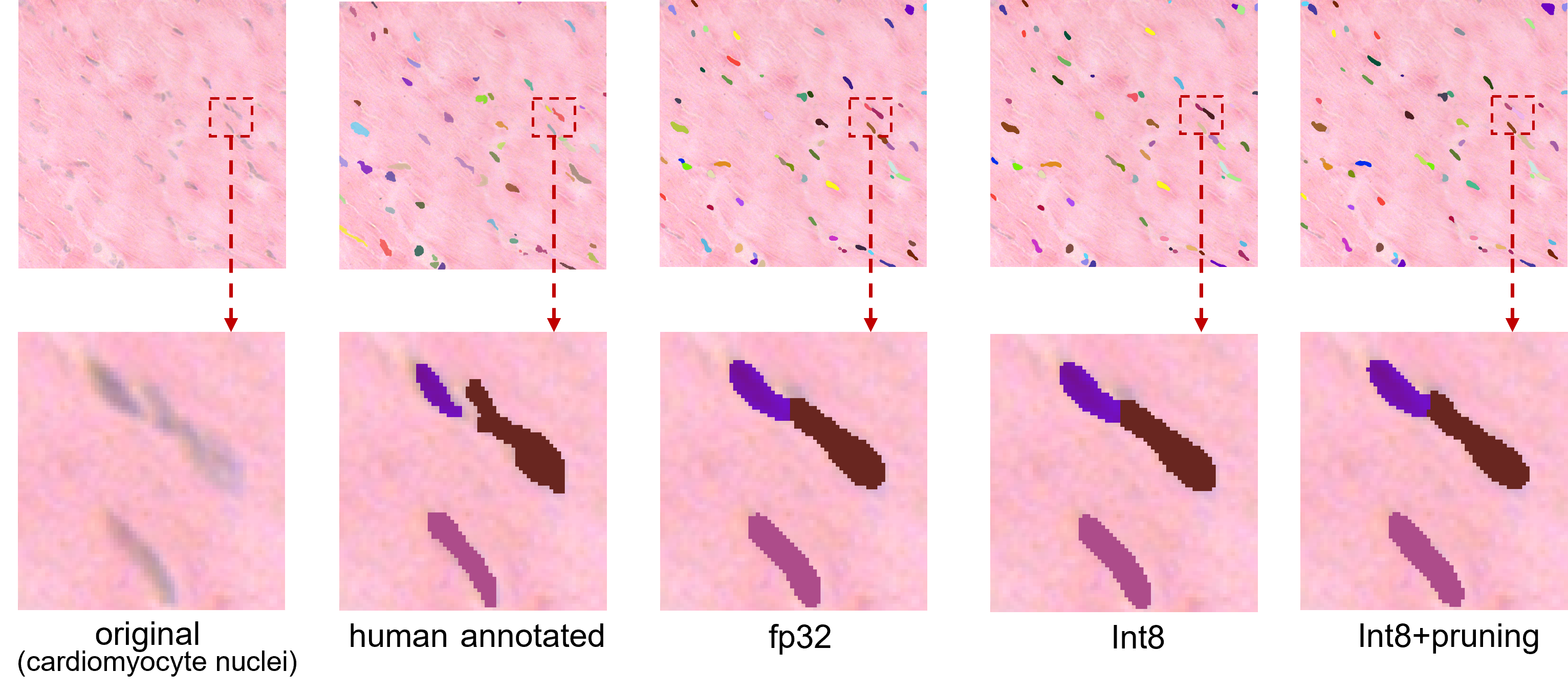}
    \caption{\textbf{Instance Segmentation Experiment Results.} There is not much difference between the inference result after compression and before compression, which implies the effectiveness of our methods.
     \textbf{Top row:} original image, human annotated labels (cardiomyocyte nuclei masks), predicted labels with float32 precision, predicted labels with integer8 precision, predicted labels with integer8 precision and pruning. \textbf{Bottom row:} The corresponding zoomed-in images.}
     \label{fig:3_instanceseg}
\end{figure}

The inference strategy is also the same, except that we change the window size to the [224,224], which is compatible with the original code. The results are shown in Fig. \ref{fig:3_instanceseg}.
    
\subsubsection*{3D osteocyte segmentation from lightsheet fluorescence microscopy images using mmv\_im2im}
    
3D segmentation is much more demanding in resource than 2D. So, we want to benchmark the efficiency improvement of bioimaging AI segmentation models on large 3d images. Here, we used the task of semantic segmentation of osteocyte from light-sheet fluorescence microscopy (LSFM) of optically cleared murine long bones as an example. The image size is around (297 $\times$ 2048 $\times$ 2048) voxels. The preparation and imaging details of this dataset are as follows.

Optical clearing was performed as described previously~\cite{gruneboom_network_2019}. In short, 4\% PFA fixed murine long bones were incubated in 50\%, 70\% and two times in 100\% ethanol, respectively, at room temperature for one day each. Finally, the samples were transferred to ethyl cinnamate (ECi, Sigma Aldrich, Cat.  112372-100G) and incubated at room temperature until they became completely transparent. LSFM of optically cleared murine long bones was performed using a LaVision BioTec Ultramicroscope Blaze (Miltenyi/LaVision BioTec, Bielefeld, Germany) with a supercontinuum white light laser (460-800 nm), 7 individual combinable excitation and emission filters covering 450 nm to 865 nm, an AndorNeo sCMOS Camera with a pixel size of 6.5 $\times$ 6.5 $\mu m ^2$, and 1$\times$ (NA 0.1), 4$\times$ (NA 0.35), 12$\times$ (NA 0.53) objectives with a magnification changer ranging from 0.6$\times$ to 2.5$\times$. For image acquisition, cleared samples were immersed in ECi in a quartz cuvette and excited at 560/40 nm for visualizing tissue autofluorescence and detected using a 620/60 nm band-pass emission filter. For CD31-AlexaFluor647 excitation a 630/60 nm band-pass filter and for detection a 680/30 nm band-pass filter were used. For image acquisition, a 4$\times$ magnification with a 4 $\mu m$ Z-step size was chosen. 

\begin{figure}[!tbp]
    \centering
    \includegraphics[width=\linewidth]{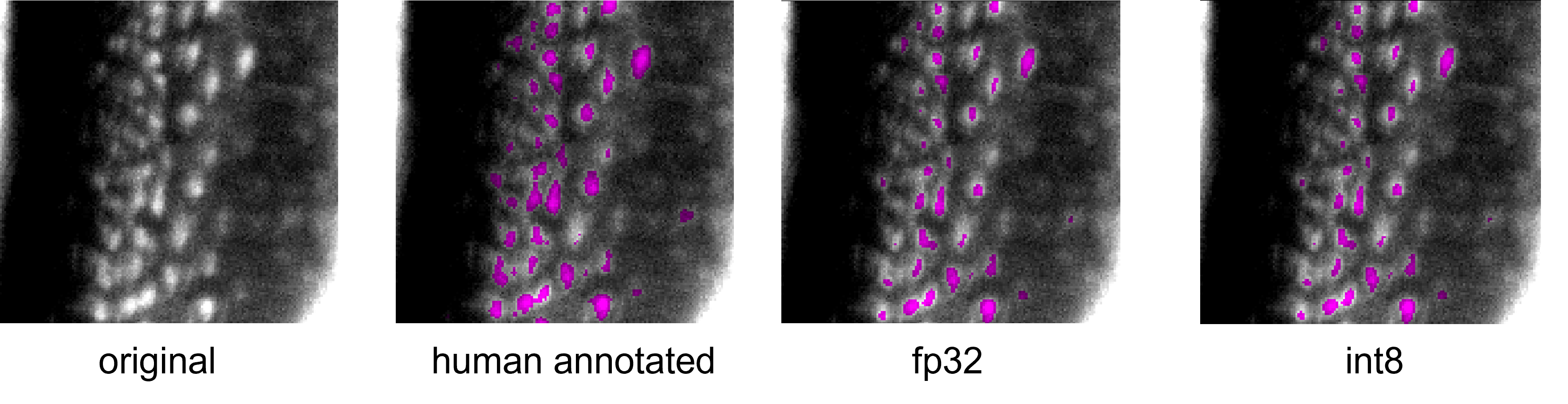}
    \caption{\textbf{Semantic Segmentation Experiment Results (a small region with human annotation available).} 
     \textbf{Top row:} original image, human annotated labels, inference output with float32 precision, inference output with integer 8 precision \textbf{Bottom row:} The corresponding zoomed-in images}
     \label{fig:3_semanticseg}
\end{figure}

The model employed was the dynamic Unet from the monai package~\cite{cardoso_monai_2022}, which characterized itself by high flexibility compared to vanilla Unet. One example patch with human annotation for evaluation was shown in Fig. \ref{fig:3_semanticseg}.

\subsubsection*{3d labelfree prediction of flurorescent images of human iPS cell from transmitted-light images}

The transmitted-light (TL) images contain rich cellular organization information but is unobtainable due to the low contrast~\cite{ounkomol_label-free_2018}. Recent models are capable of directly transforming the low-resolution TL images to the fluorescent images with clear labeled subcellular structures~\cite{ounkomol_label-free_2018,sonneck_mmv_im2im_2023}, without the requirement for fluorescent labeling. 

\begin{figure}[!tbp]
    \centering
    \includegraphics[width=\linewidth]{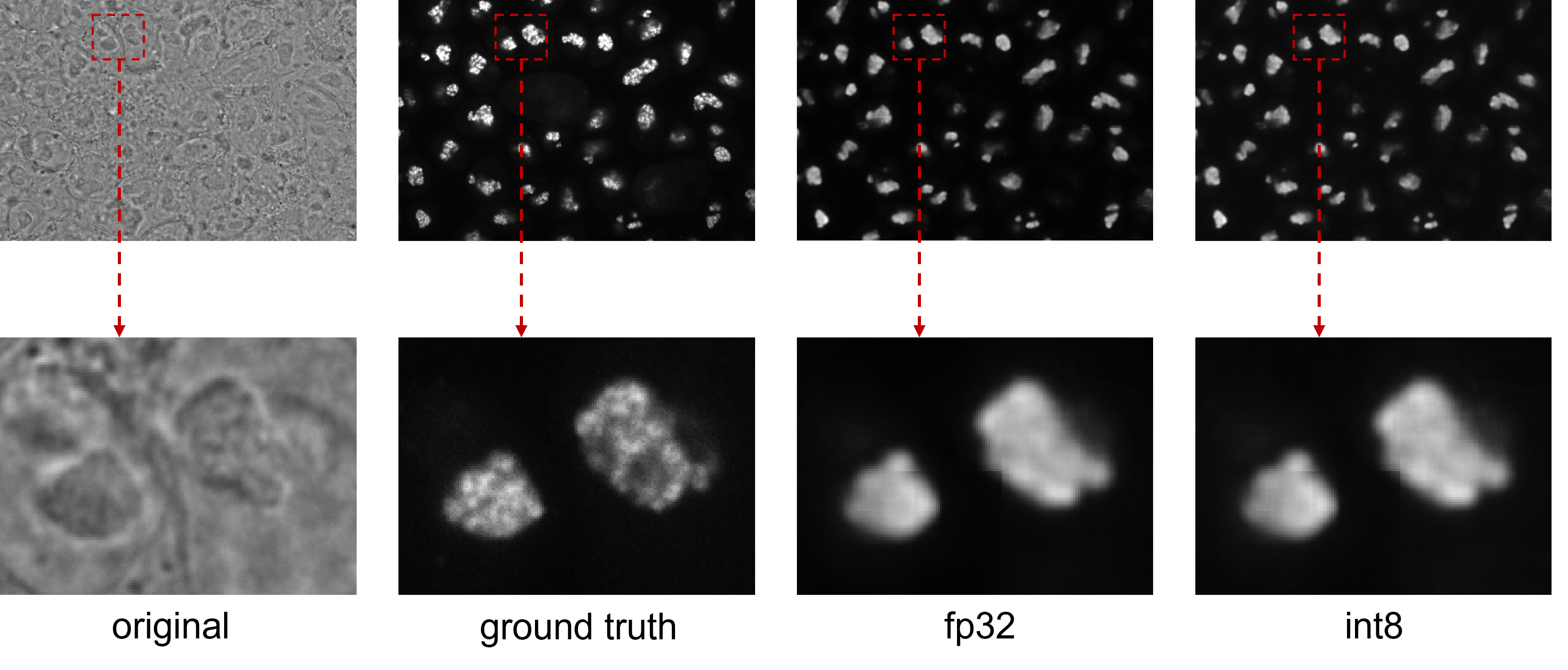}
    \caption{\textbf{Labelfree Experiment Results.} There is not much difference between the inference result after compression and before compression, which implies the effectiveness of our methods.\textbf{Top row:} original image, ground truth, inference output with float32 precision, inference output with integer 8 precision \textbf{Bottom row:} The corresponding zoomed-in images}
    \label{fig:3_labelfree}
\end{figure}

The dataset selected here was from the public dataset released with~\cite{viana_integrated_2023} and we chose mmv\_im2im fnet 3D as our model and the fibrillarin cellline for the purpose of demonstration. The compression and inference part were the same as the denoising task. The results were shown in Fig. \ref{fig:3_labelfree}. It is important to emphasize that the focus of this work is not to achieve the best labelfree prediction. The most important evaluation is to make sure the prediction from the compressed model is sufficiently similar to the prediction from the original model before compression.

\end{document}